# Building High-Quality Auction Fraud Dataset

Sulaf Elshaar[1] & Samira Sadaoui[1]

[1] Department of Computer Science, University of Regina, Regina, Canada

Correspondence: Sulaf Elshaar, Faculty of Graduate Studies and Research, Department of Computer Science, University of Regina, Regina, Canada.

**Abstract**

Given the magnitude of online auction transactions, it is difficult to safeguard consumers from dishonest sellers, such as shill bidders. To date, the application of Machine Learning Techniques (MLTs) to auction fraud has been limited, unlike their applications for combatting other types of fraud. Shill Bidding (SB) is a severe auction fraud, which is driven by modern-day technologies and clever scammers. The difficulty of identifying the behavior of sophisticated fraudsters and the unavailability of training datasets hinder the research on SB detection. In this study, we developed a high-quality SB dataset. To do so, first, we crawled and preprocessed a large number of commercial auctions and bidders' history as well. We thoroughly preprocessed both datasets to make them usable for the computation of the SB metrics. Nevertheless, this operation requires a deep understanding of the behavior of auctions and bidders. Second, we introduced two new SB patterns and implemented other existing SB patterns. Finally, we removed outliers to improve the quality of training SB data.

## 1. Introduction

Unquestionably, e-commerce has taken the world by storm. In 2017, this sector accounted for over 2.3 trillion dollars in sales with an expected increase to 4.5 trillion by 2021 (statista.com nd). Several factors drive up the demand for e-commerce transactions, including the 24/7 accessibility, product availability and variety, detailed product description, and friendly return policies. One part of this economic activity is the lucrative e-auction marketplace. Given the magnitude of the monetary transactions, auction sites are very attractive to fraudsters and scam artists. According to the Internet Crime Complaint Center (IC3), in the year 2015 alone, 21 510 complaints related to auction fraud have been recorded along with a financial loss estimated to $19 million (IC3, 2015). Fraudulent activities of e-auctions typically occur within three-time frames: pre-auction (ex. misrepresentation of items and selling of black-market items), in-auction (ex. shill bidding and bid shielding) and post-auction (ex. non-delivery of items) (Ford, Xu & Valova, 2012). Both pre- and post-auction fraud can be documented by participants because in most cases they are supported by physical evidence. However, in-auction fraud, which happens during the bidding period, does not leave any obvious evidence, and worst of all it is often unnoticed by the victims i.e. the winning buyers (Xu, Bates & Shatz, 2009). Shill Bidding (SB) is a pernicious online activity that plagues the auction industry. Shill bidders are merely shills or fakes who compete in consort with accomplice sellers. Their sole purpose is to artificially elevate the prices for the items being auctioned. To date, there are no solid statistics that measured the financial impact caused by this type of fraud, and for the most part, the annual number of victims and the magnitude of their losses are unknown. Yet, the eBay community webpage reveals numerous anecdotal complaints from genuine buyers and sellers together with their monetary losses. SB fraud may undermine the confidence of bidders, buyers and even sellers as explained below:

·       *Bidders* attempt to detect SB by themselves by tracking many of the competitor's behavior and communicating their suspicions to eBay. Very recently, the bidders' IDs and history are no longer available on eBay. We believe this new policy about blocking the bidding history is to not be able to discover SB activities.

·       *Buyers* are the most affected by SB since they pay much more for the items. The price is driven up by disingenuous bidders with no intention of ever winning the bid. For instance, CNB News disclosed that a bidder paid $1, 825 for a nearly complete set of 1959 Topps baseball cards on eBay (nbcnews.com nd). However, two undercover detectives determined that the purchaser ended up paying an extra $531 for the cards due to SB conducted by the seller with an alternate identity.

- *Sellers* complain that they are wrongly accused of SB by eBay. Several sellers reported having received warning messages from eBay that SB has been found in their auctions after which their auctions are removed and their fees forfeited. They complained that they often lose the opportunity to participate in future auctions while other sellers who do actually engage in SB go unrecognized and unpunished. Indeed, to avoid being detected, a shill may mimic normal bidding behavior, or use several phony accounts, or hire fake bidders anywhere in the world to participate in the scam. Given the large volume of auctions conducted every day in a single commercial site like eBay, it is very challenging to monitor the behavior of bidders in order to detect in real-time SB fraud, especially when the auctions involve many users and have long durations, like 7 and 10 days. Machine Learning Algorithms (MLAs) represent a vital solution to address this challenge. To date, the application of MLAs to auction fraud has been restricted, unlike other fraud perpetrated in other industrial sectors, such as credit card, telecommunication, and insurance. In addition to the difficulty of identifying appropriate SB patterns, researchers are hindered by the absence of training data that are used to develop SB classification models. The availability of authentic data, denoting the real behavior of bidders, is essential for building robust fraud detection models. Nevertheless, obtaining SB data is a very laborious task as demonstrated in this study. Therefore, our main objective is to build a high-quality fraud dataset for the e-commerce auctions based on the most recent real-world data and then share it with the ML community to conduct more research in the context of online auction fraud detection.

In fact, we need first to scrape a large number of auctions from a commercial website, preprocess the raw data, implement algorithms to measure the SB patterns, and finally evaluate each SB metric against each bidder in each collected auction in order to produce the SB training dataset. In the following, we highlight the main contributions of our research when compared to past studies:

1. Our empirical data analysis employs the most recent fraud data. In fact, from the eBay website, we crawled a good number of auctions of a very hot product as well as the bidder history. Note this is the first time the bidder history has been utilized to measure the SB patterns. We may mention that the preprocessing of the two raw datasets is very difficult and requires a deep understanding of the bidding behavior. We aim to share the auction and bidder history datasets as well as the SB training dataset with the research community.

2. After an extensive examination of the eBay policies as well as the scraped auction and bidder data, we introduced two new SB strategies and defined algorithms to measure them. Moreover, we improved and implemented the metrics of seven other SB patterns from the literature.

## 2. Scraping of Auctions and Bidder History

We crawled data from the eBay website for a period of three months, March 31 to June 28, 2017. To do so, we employed a professional Web scraper, called Octoparse, which retrieved auctions of a hot product based on user-defined filters. From the URL page containing the auction listing, each product page is parsed and information is then extracted. We collected Forward, English auctions of the "iPhone 7" by applying three filters:

a) North America products on eBay.com,

b) Cellphones and Accessories category, and c) the winning price of an auction must be more than $100. Moreover, for each bidder that participated in the targeted auctions, we scraped their full history information. We note that we did not extract proxy bids that eBay automatically places on behalf of users because our focus is on the actual bidding behavior.

We utilized a tool called Terapeak to get the statistics on the iPhone7 sales in eBay during the period in which we extracted the data. We then compared the sales of our product in several categories. In the period of May-June 2017, iPhone7 sales belonged to the topmost sold category. Also, it has been cited as a hot product on eBay.com during the same period, which means it is among the most sold items. Moreover, 93% of iPhone7 sales belong to the Cellphone and Accessories category, hence our decision to study eBay sales in this category. We selected iPhone 7 for reasons that may have likely encouraged SB activities:

- It attracted a large number of bidders and bids, so its auctions are interesting to examine.

- The winning prices are high, in our raw data, the average value is $565.3. We note that 34 % of winning bids are above the average. More the price is high, more the possibility of SB (Dong, Shatz & Xu, 2009).

- The bidding duration varies between 1, 3, 5, 7 and 10 days. The longer the duration, the more the

chance for a shill bidder to imitate normal bidding behavior (Dong, Shatz & Xu, 2009).

The two collected raw datasets are exposed in Table 1. The auction dataset consists of information about auctions, bidders, and sellers. The bidder history dataset contains the activities of each participant in the last 30 days.

Table 1. Statistics of Raw iPhone 7 Data

|  | Auction Dataset | Bidder History Dataset |
|---|---|---|
| Number of Auctions | 2551 | NA |
| Number of Bidder IDs | 1226 | 6523 |
| Number of Records | 399206 | 404239 |
| Number of Attributes | 28 | 15 |
| Number of Records with Foreign Currency | 446 (113 CAN, 71 GBP, 263 EUR) | NA |

## 3. Raw Data Preprocessing

We thoroughly preprocessed both datasets to make them usable for the computation of the SB metrics. Nevertheless, this operation requires a deep understanding of the behavior of auctions and bidders. The higher the quality of the training data, the more reliable the classification outcome.

*3.1 Cleansing Data*

It is not surprising that real-world data are imperfect. Both datasets possess missing values as well as duplicated records. These defects happen during data extraction from websites. In fact, records are duplicated when acquiring information from a link that leads to several other links containing the same information. Hence, in the first stage, we cleaned up the raw datasets as follows: 1) we deleted duplicated records from Auction and Bidder History tables; 2) we removed records with missing or fully masked bidder' IDs. Although the bidder ID is a significant attribute, we deleted those records because we cannot impute the value for an ID, and 3) we eliminated attributes that are irrelevant to our SB patterns.

*3.2 Reformatting and Merging Data*

Another problem we encountered was the improper format of the attributes of interest. In Figure 1, we present the transformation process of four attributes related to the date and time. We first reformatted these attributes into their equivalent serial numbers and then merged them into a single attribute called "Bid Time" that denotes the elapsed time between the auction starting time and the placed bid time. As an example, in Figure 1, Bid Time of 1.2862 means that the bid has been submitted after one day and 0.2862 since the auction begun (times are stored as fractions of a full day).

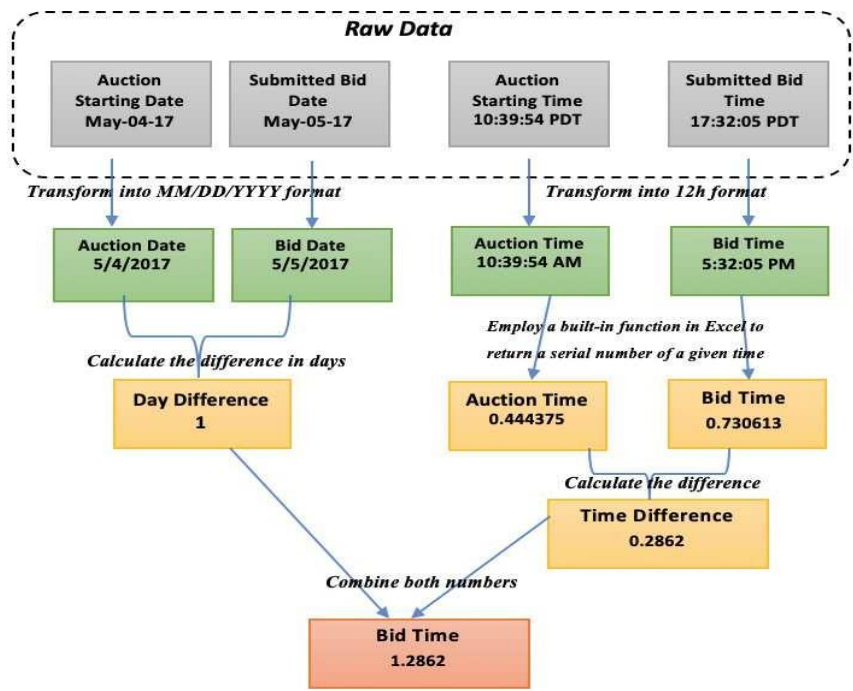

Figure 1. Reformatting and Merging Attributes

Another issue is the multiplicity of currency in addition to the US dollar, such as the Canadian dollar, British Pound and European Euro. Although they represent only 1.5 % of the dataset, they cannot be ignored. Hence, using the currency conversion of Kutools, an add-in tool in Excel, we converted the currency of two attributes, "Bid Amount" and " Opening Price", into US dollar. We utilized the currency exchange rates w. r. t. the period in which the data has been collected. Furthermore, in the Bidder History table, the activity with a specific seller has been given in a percentage, so we transformed it into a number (e.g. 90% becomes 0.90). We also changed the bidding duration from a textual to a numeric format (e.g. "7 Days" becomes 7).

*3.3 Maintaining Data*

For the identification purpose, we included a new attribute named "Auction ID". For this purpose, we generated a unique integer identifier for each auction based on two attributes: "Seller ID" and "Product URL". As an example, an auction ID is obtained from ***abcde*** and

eBay.com/itm/Apple-iPhone-7-Plus-PRODUCT-RED-256GB-Unlocked-S03tphone-New-With-Freebies-/15250 6042803.

There are some attributes that possess inconsistent values that will lead to deceptive performance results if not corrected. For example, the Winning Price attribute does not match the highest bid in several auctions, and the Number of Bids attribute is often more than the number of actually submitted bids. In both cases, we replaced the invalid values with the corrected ones. Table 2 exposes the statistics after preprocessing the auction and bidder history datasets.

Table 2. Statistics of Preprocessed iPhone 7 Data

|  | Auction Dataset | Bidder History Dataset |
| --- | --- | --- |
| Number of Auction IDs | 1444 | NA |
| Number of Bidder IDs | 1163 | 1230 |
| Number of Records | 29000 | 8853 |
| Number of Attributes | 9 | 5 |
| Average of Winning Price | 641.24 | NA |
| Average of Starting Price | 141.33 | NA |
| Average of Number of Bids | 60.59 | NA |

**4. SB Strategies**

*4.1 Existing SB Strategies*

The prior work (Sadaoui & Wang, 2017) investigated the literature and compiled several SB patterns that have been found to be frequent in auctions infected with SB. The description of these patterns is given in Table 3. We assign different weights to the fraud patterns w. r. t. their behavior and importance, which are divided into three scores: Low, Medium and High. For instance, we give "Early Bidding" a low weight because a normal bidder can still bid early. "Winning Ratio" has a high weight because it is a strong indicator of fraud: a shill bidder has high participation in his history, yet he rarely wins. We give "Buyer Tendency" a medium weight because there is a chance that the bidder is very interested in the product, or he prefers to buy from a reputable seller a good reputation. We implemented the seven patterns in the MS-SQL server 2017 based on the formulas given in (Sadaoui & Wang, 2017). The value of an SB metric is in the range of [0, 1]; the higher the value, the more suspicious the bidder being examined.

Table 3. Existing SB Strategies

| Name | Description | Weight |
| --- | --- | --- |
| Auction Opening Price | Auctions with a low opening price are more likely to in volve SB | Low |
| Early Bidding | Shills start bidding very early to attract the attention of other users | Low |
| Last Bidding | Shills do not place bids in the last period of an auction to avoid winning | Medium |
| Bidding Ratio | Shills compete in an auction much more than normal bid- ders to inflate the price | Medium |
| Auction Bids | Auctions with shilling have often more bids than concur- rent auctions (selling the same product in the same time period) | Low |
| Buyer Tendency | A shill participates in auctions of a particular seller more than other sellers with the same product | Medium |
| Winning Ratio | Shills avoid winning despite their large number of bids | High |

We also investigated "Bid Increment" pattern (Sadaoui & Wang, 2017); (Tsang, Koh, Dobbie, & Alam, 2014) where bidder bids with a very low increment. Although this fraud pattern has been used in several studies, it is inappropriate for our extracted auction data because in several cases a bid amount is not higher than the previous one. For example, as shown in Figure 2, the user g***a bided with $312, and just after that, another user n***8 bided less with $250. However, in the English auction protocol, the bid amounts must be ascending. So, we manually checked whether these cases are errors from eBay itself or due to the data scraping operation. We found out that many auctions show that these faults occur in the eBay site, e.g. in Figure 3, $581 was the highest bid submitted at 5: 45:59 PM, and after 5 hours, the user i***e placed two lower amounts of $571 and $541.00.

Figure 2. Bid Amount in our Auction Dataset

Figure 3. Bid Amount in an Auction of January 2018

*4.2 New SB Strategies*

After examining thoroughly, the eBay rules and policies, we introduced in this study two new fraud patterns. On eBay, sellers can rate buyers and vice versa but only after the transactions have been completed in order to make the feedback more reliable. As stated in the eBay webpage (2013): *"Few shill bidders will have any feedback. This is because they never follow through with transactions and therefore do not have anyone to leave feedback for them"*. In the past research (Ford, Xu, & Valova, 2012); (Dong, Shatz, Xu, & Majumdar,2012), the "Buyer Rating" (i.e. the number of feedbacks) has been utilized as an individual SB pattern while in our work, we use it together with the number of products a bidder bided on in the past 30 days. The new pattern, called "Buyer Rating Based on Items", takes into consideration the following points (see Algorithm 1):

· A high bidder rating means the bidder proceeded into many transactions. This indicates a usual bidding behavior.

· A genuine bidder may take a long time to build a good reputation. In this case, he would care about his reputation and avoid using his real account for a shilling.

· A shill bidder usually opens a fake account to commit fraudulent activities.

· Not every zero rating is a sign of SB. We need to compare it with the number of items auctioned in the previous month. Here, we suggest that if a bidder bided on more than five items in the last month and has no rating, this may indicate suspicious behavior because that majority of bidders (∼ 96.3%) with five bids have gotten at least one rating according to our bid history data.

· If the bidder rating is greater than or equal to the number of items the user bided on, this indicates he has a good reputation and used his real account. Otherwise, we calculate the ratio between the bidder rating and the number of items that the user bided on. The lower the ratio, the more likely the user is a shill.

Algorithm 1: Buyer Rating Based on Items (BRBI)
1 Inputs: BuyerRating, ItemsBidOn30Days
2 $BRBI \leftarrow 0$
3 if $BuyerRating = 0$ then
4     if $ItemsBidOn30Days >= 5$ then
5         $BRBI \leftarrow 1$
6     else if $BuyerRating < ItemsBidOn30Days$ then
7         $BRBI \leftarrow BuyerRating/ItemsBidOn30Days$

As examples, in Table 4, the user a***e bided on only one item and got zero feedback. This indicates a normal case. The bidder z***c bided on eight items and has no rating, so he is probably a fraud. We tracked this user manually in the bidder history and found out that 30% of his auctions were for one seller only, and he retracted a quarter of his auctions for the same seller. The bidder s***s obtained 2715 feedbacks even though he participated in only one auction during the last month. This means his account is older than one month and he proceeded to transactions a lot. On the contrary, the user w***w have fewer feedbacks than the number of items bid in the previous month. So, we calculated the ratio that has a low value of 0.23333, which shows that he is most likely a normal bidder.

Table 4. Buyer Rating Compared to Number of Items Bid in Last 30 Days

| Bidder ID | Bidder Rating | Items Bid on 30 Days | BRBI |
|---|---|---|---|
| a***e | 0 | 1 | 0 |
| z***c | 0 | 8 | 1 |
| s***s | 2715 | 1 | 0 |
| w***w | 7 | 30 | 0.23333 |

According to the eBay website, the bid retraction can be a sign of SB that can result in the suspension of the users' account. Bids can be retracted under two restrictions, a valid reason and the time of retraction. The reasons are threefold:

1) the bidder typed in a wrong bid amount, e.g. he bided $777.50 instead of $77.50; 2) the item description changed after he entered his last bid; 3) he cannot contact the seller by phone or email. If the bidder meets one of the above conditions, he should fill out the bid retraction form. After that, he should meet certain time restrictions. We chose to use "Bid Retraction" as a fraud pattern because eBay explicitly states it is a strong sign of SB, especially within the last 24 hours. Since we do not have the time of retraction in our data (only the number of retractions during the last 30 days), we employ the number of retractions of a bidder along with his activity with a specific seller. The activity with a seller denotes the percentage of the total number of bids placed by a bidder for a specific seller during the last month. When *ActivityWithSeller* = 1, it means the activity of a bidder with a certain seller is %100. We consider that if a bidder retracts bids and has a very strong relationship with a seller, then this indicates a colluding behavior. According to our bidder history data, the average of the bidders do not retract their bids, and around 7% of the bidders retract more than the average. The algorithm to compute this pattern from the bidder history is exposed in Algorithm 2. "Buyer Rating based on Items" has low weighed because buyers may not get ratings from sellers even if they have proceeded to transactions for several products. "Bid Retraction" is a strong sign of SB in eBay and since we used with the relation with a specific seller if the number of retractions is small, so we give it medium weight. We implemented 1 and 2 in the *VBA- MS Visual Basic for Applications*.

---

Algorithm 2: Bid Retraction BR

1　Inputs: NoOfBidRetractions, ActivityWithSeller
2　$BR \leftarrow 0.5$
3　if (*NoOf BidRetractions* >= 1) & (*ActivityWithSeller* = 1) then
4　　　$BR \leftarrow 1$
5　else if (*NoOf BidRetractions* >= 1 ) & (*ActivityWithSeller* >= 0.7) then
6　　　$BR \leftarrow 0.5$

---

## 5. Outlier Filtering

To build the training dataset, we measured the values of the nine SB patterns against each bidder in each of the 1444 auctions. As a result, we produced a tally of 11 954 SB samples. Each sample, which represents the bidder's conduct in a certain auction, is a vector consisting of the Auction ID, Bidder ID, and values of the nine fraud patterns.

Before the data labeling task, we need first to check our SB dataset for outliers since they cause dire effects on the classification accuracy (Brownlee, 2016). Outliers denote samples that are not compatible with the rest of the dataset. We employ the Inter Quartile Range (IQR), one of the outlier labeling methods originally introduced in (Tukey, 1977). IQR uses a box plot to display information about continuous univariate data, such as the median, lower quartile, upper quartile, lower extreme and upper extreme values of a dataset. We chose *IQR* to detect outliers owing to the following factors:

·　　Based on the sample mean and standard variance, IQR is less vulnerable to extreme values than other filtering methods because it uses quartiles that are resistant to extreme values (Walfish, 2006); (Barbato, Barini, Genta & Levi, 2011).

- It can be employed for any data distribution since it depends on the median and not the mean (Walfish, 2006); (Barbato et al., 2011).
- It is simple to apply and there are many supporting statistical tools.

After identifying outliers in our training dataset, we treated them in two different ways: extreme values that are out of the range and within the range. First, we discarded all the values that are out of the range of [0,1]. These unwanted values, found in the "Early Bidding" and "Last Bidding" features, are after the auction ends or before the auction starts. As an example, in Figure 4, we can see the extreme values, in the range of [1, 7], lie above the upper limit of the Box Plot for the "Last Bidding". Second, for the values that in the range, we decided to normalize them because some attribute distributions are not Gaussian. Our goal is to prepare the SB dataset to be used for different MLAs where some of them (e.g. Support Vector Machines, K-Nearest Neighbors and K-Means) require data rescaling. For instance, Figure 5 depicts the distribution of "Winning Ratio" values before and after normalization. Table 5 presents the SB dataset before and after the outlier filtering.

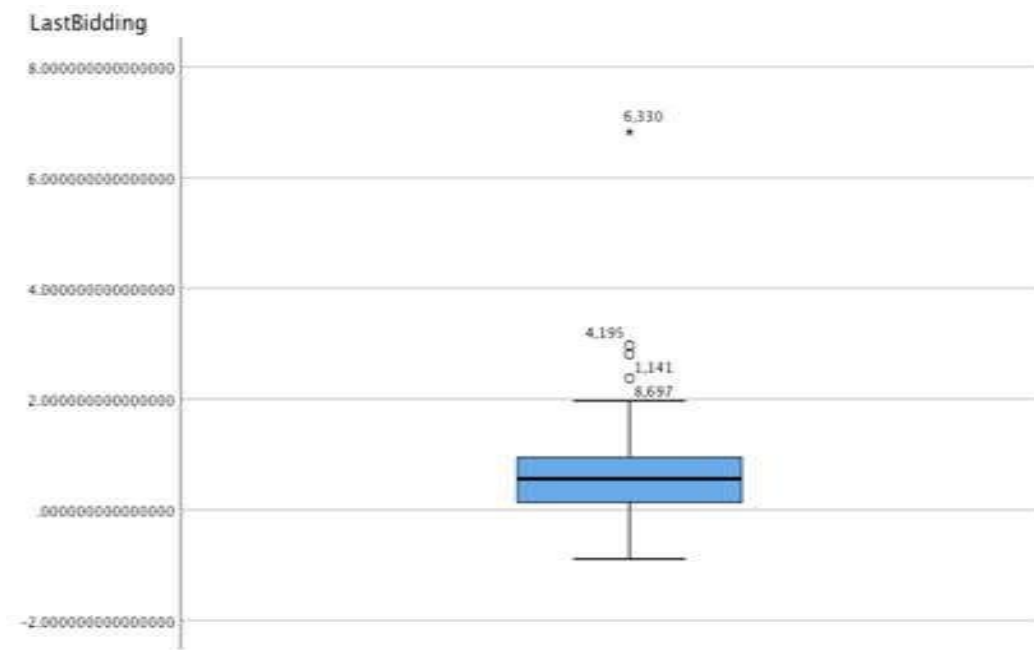

Figure 4. Scatter Graph of Last Bidding

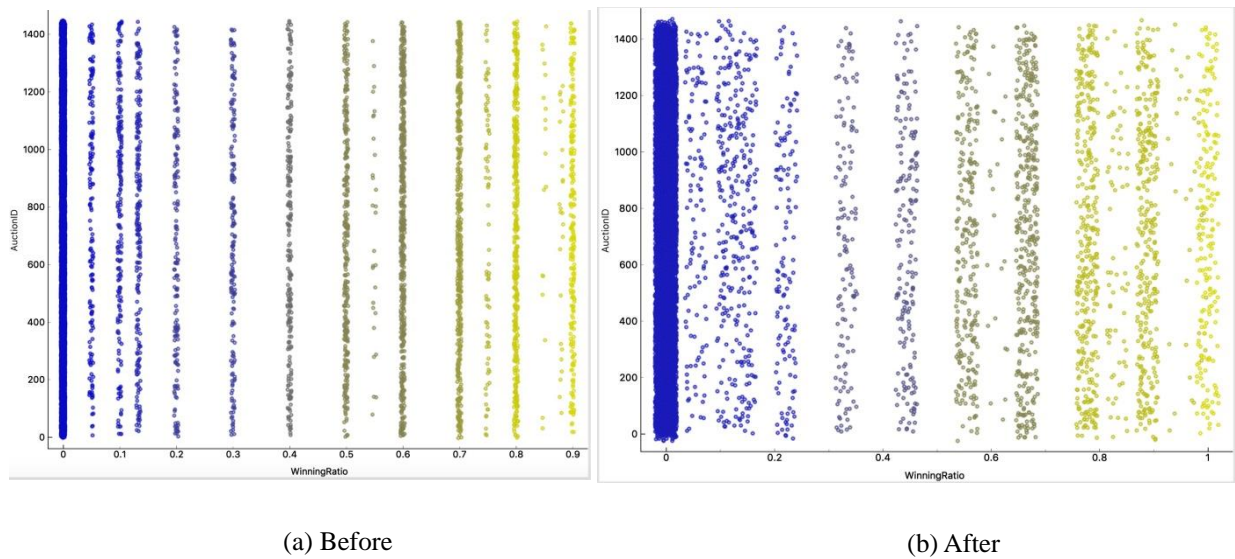

(a) Before  (b) After

Figure 5. Scatter Graph of Winning Ratio Before and After Outliers Detection

Table 5. SB Dataset Before and After Outlier Filtering

|  | Before | After |
| --- | --- | --- |
| Number of Auction IDs | 1444 | 1399 |
| Number of Bidder IDs | 1163 | 1100 |
| Number of Records | 11954 | 9291 |

## 6. Conclusion

SB is a serious cyber-crime driven by the capabilities of modern-day technologies and clever scammers. Hence, robust detection methods must be developed to effectively identify SB activities. The empirical analysis of bidding fraud data has been limited in the literature due to the difficulty in identifying SB strategies on one hand, and the non-availability of training SB data on the other hand. In this study, we have built a high-quality SB training dataset that can be used by different machine learning methods. To do so, first, we have extracted a large number of data from commercial auctions and bidders' history as well. Subsequently, we have conducted several time-consuming operations to preprocess the raw data. For the first time, we have introduced two new SB patterns," Bid Retraction"and" Buyer Rating based on Items".

## 7. Future Work

The next phase of this research is to label the SB instances into Normal or Suspicious, which is a very challenging task. To this end, we would like to apply the Semi-Supervised Classification (SSC) approach where only a few labeled data are needed. In SSC, the model can learn efficiently from a few labeled data along with many unlabeled data. To the best of our knowledge, this will be the first attempt to employ SCC for SB detection. For this purpose, we need first to label a portion of our SB dataset, for instance through data clustering techniques (Alzahrani & Sadaoui, 2019). Moreover, we will study how to deal with the problem of imbalanced class distribution, which is expected in fraud classification applications (Ganguly & Sadaoui, 2017); (Anowar, Sadaoui & Mouhoub, 2018).